\documentclass[11pt,reqno]{amsart}
\usepackage{graphicx,verbatim,lineno,titletoc}
\usepackage{amssymb,mathrsfs}
\usepackage{amsmath}
\usepackage{adjustbox}
\usepackage{calc}
\usepackage{ytableau}
\usepackage{array}
\usepackage{tikz}
\usetikzlibrary{shapes.multipart,patterns,arrows}
\usepackage[font=footnotesize]{caption}
\usepackage[T1]{fontenc} 
\usepackage{fourier} 
\usepackage[english]{babel} 
\usepackage{appendix}
\usepackage[all]{xy}
\usepackage{enumerate}
\usepackage[english]{babel}
\usepackage{multirow}
\usepackage{float}
\usepackage{enumitem}
\usepackage{accents}
\usepackage[numbers]{natbib}
\usepackage{hyperref}
\hypersetup{colorlinks}
\usepackage{algorithm}
\usepackage[noend]{algpseudocode}
\usepackage{multicol}

\usepackage[top=1in, left=1.0in, right=1.0in, bottom=1in]{geometry}
\hypersetup{colorlinks,linkcolor={red},citecolor={olive},urlcolor={red}}

\newcolumntype{M}[1]{>{\centering\arraybackslash}m{#1}}
\newcolumntype{N}{@{}m{0pt}@{}}

\pdfoutput=1

\DeclareMathOperator*{\argmin}{arg\,min}

\theoremstyle{definition}

\allowdisplaybreaks

\makeatletter
\newcommand{\addresseshere}{%
    \enddoc@text\let\enddoc@text\relax
}
\makeatother

\begin{document}

\title[Topic-aware Chatbot Using RNN and NMF]{Topic-aware Chatbot Using Recurrent Neural Networks \\ and Nonnegative Matrix Factorization}

\author[Y. Guo]{Yuchen Guo}
\address{Yuchen Guo, Hunan University, Changsha, Hunan, 410082, China}
\email{\texttt{guoyuchen1228@vip.qq.com}}

\author[N. Hanoian]{Nicholas Hanoian}
\address{Nicholas Hanoian, University of Vermont, Burlington, VT 05405, USA}
\email{\texttt{nhanoian@uvm.edu}}

\author[Z. Lin]{Zhexiao Lin}
\address{Zhexiao Lin, Zhejiang University, Hangzhou, Zhejiang, 310058, China}
\email{\texttt{zhexiaolin0926@gmail.com}}

\author[N. Liskij]{Nicholas Liskij}
\address{Nicholas Liskij, Department of Mathematics, University of California, Los Angeles, CA 90095, USA}
\email{\texttt{nicholas.liskij@gmail.com}}

\author[H. Lyu]{Hanbaek Lyu}
\address{Hanbaek Lyu, Department of Mathematics, University of California, Los Angeles, CA 90095, USA}
\email{\texttt{hlyu@math.ucla.edu}}

\author[D. Needell]{Deanna Needell}
\address{Deanna Needell, Department of Mathematics, University of California, Los Angeles, CA 90095, USA}
\email{\texttt{deanna@math.ucla.edu}}

\author[J. Qu]{Jiahao Qu}
\address{Jiahao Qu, Department of Mathematics, University of California, Los Angeles, CA 90095, USA}
\email{\texttt{jiahaoucla@gmail.com }}

\author[H. Sojico]{Henry Sojico}
\address{Henry Sojico, Harvey Mudd College, Claremount, CA 91711, USA
}
\email{\texttt{hsojico@hmc.edu}}

\author[Y. Wang]{Yuliang Wang}
\address{Yuliang Wang, Shanghai Jiao Tong University, Shanghai, 200240, China
}
\email{\texttt{2215429764@qq.com}}

\author[Z. Xiong]{Zhe Xiong}
\address{Zhe Xiong, Shanghai Jiao Tong University, Shanghai, 200240, China 
}
\email{\texttt{aristotle-x@sjtu.edu.cn}}

\author[Z. Zou]{Zhenhong Zou}
\address{Zhenhong Zou, Beihang University, Beijing, 100083, China
}
\email{\texttt{joebuaa2016@gmail.com}}


\begin{abstract}
We propose a novel model for a topic-aware chatbot by combining the traditional Recurrent Neural Network (RNN) encoder-decoder model with a topic attention layer based on Nonnegative Matrix Factorization (NMF). After learning topic vectors from an auxiliary text corpus via NMF, the decoder is trained so that it is more likely to sample response words from the most correlated topic vectors. One of the main advantages in our architecture is that the user can easily switch the NMF-learned topic vectors so that the chatbot obtains desired topic-awareness. We demonstrate our model by training on a single conversational data set which is then augmented with topic matrices learned from different auxiliary data sets. We show that our topic-aware chatbot not only outperforms the non-topic counterpart, but also that each topic-aware model qualitatively and contextually gives the most relevant answer depending on the topic of question. 
\end{abstract}

${}$
\vspace{-0.5cm}
${}$
\maketitle
 \begin{multicols}{2}

\section{Introduction}
\label{Introduction}

Recently, deep learning algorithms \cite{deng2013recent,bengio2013deep,deng2014tutorial} have demonstrated significant advancements in various areas of machine learning including image classification \cite{krizhevsky2012imagenet},  computer vision \cite{boureau2010theoretical}, and voice recognition tasks \cite{hannun2014deep,amodei2016end}, even outperforming hand selected features from experts with decades of experience. 

Another area where deep learning algorithms have been successfully applied is \textit{sequence learning}, which aims at understanding the structure of sequential data such as language, musical notes, and videos. One example of an application of deep learning in language modeling is conversational \textit{chatbots}. A chatbot is a program that conducts a conversation with a user by simulating one side of it. Chatbots receive inputs from a user one message, or question, at a time, and then form a response that is sent back to the user. One of the most widely used machine learning techniques for sequence learning is \textit{Recurrent Neural Networks} (RNN). In the viewpoint of statistical learning theory and generative models, sequence learning can be regarded as the problem of learning a joint probability distribution induced by a given sequence data corpus. RNNs learn such joint probability distributions by learning all conditional probability distributions through a deep learning architecture. 

A complementary approach in machine learning is \textit{topic modeling} (or \textit{dictionary learning}), which aims at extracting important features of a complex dataset so that one can represent the dataset in terms of a reduced number of extracted features, or topics. One of the advantages of topic modeling-based approaches is that the extracted topics are often directly interpretable, as opposed to the arbitrary abstractions of deep neural networks. Topic models have been shown to efficiently capture intrinsic structures of text data in natural language processing tasks \cite{steyvers2007probabilistic, blei2010probabilistic}. Two prominent methods of topic modeling are \textit{Latent Dirichlet allocation} (LDA) \cite{Blei2003} and \textit{nonnegative matrix factorization} (NMF) \cite{lee1999learning}, which are based on Bayesian inference and optimization, respectively. 


An active area of research in machine learning is combining the generative power of deep learning algorithms with the interpretability of topic modeling. Recently, new algorithms combining deep neural networks with LDA or NMF have been proposed and studied, aiming at better performance and human interpretability in various classification tasks \cite{trigeorgis2016deep, flenner2017deep, jin2018combining, MNW19Bias}. 

We are interested in combining RNN-based chatbot models with topic models to enable chatbots to be aware of the topics of input questions and give more topic-oriented and context-sensible output. In fact, there are recent works which combine RNN-based chatbot architecture with LDA-powered topic modeling \cite{xing2017topic, serban2017hierarchical}. However, there has not yet been a direct attempt to incorporate the alternative topic modeling method of NMF with chatbots. Using NMF over LDA provide many advantages including computational efficiency, ease of implementation, possibility of transfer learning, as well as the possibility of incorporating recent developments of online NMF on Markovian data \cite{lyu2019online}.

\subsection{Our contribution}

In this paper, we construct a model for a topic-aware chatbot by combining the traditional RNN encoder-decoder model with a novel topic attention layer based on NMF. Namely, after learning topic vectors from an auxiliary text corpus via NMF, each input question that is fed into the encoder is augmented with the topic vectors as well as its correlation with topic vectors. The decoder is trained so that it is more likely to sample words from the most correlated topic vectors.

We demonstrate our proposed RNN-NMF chatbot architecture by training on the same conversation data set (Cornell Movie Dialogue) but supplemented with different topic matrices learned from other corpora such as Delta Airline customer service records, Shakespeare plays, and 20 News groups articles. We show that our topic-aware chatbot not only outperforms the non-topic counterpart, but also each topic-aware model qualitatively and contextually gives the most relevant answer depending on the topic of question.

\section{Background and Related Works}

There are two main approaches for building chatbots: retrieval-based methods and generative methods. A retrieval-based chatbot is one that has a predefined set of responses, and each time a question is asked, one of these pre-written responses will be returned. The advantage here is that every response is guaranteed to be fluent. Many researchers have come up with interesting ideas to build a chatbot that can extract the most important information from a question and choose the correct answer from a given set of possible responses \cite{wu2016sequential,wang2013dataset}. Generative chatbots, on the other hand, have the ability to generate a new response to each given input question. These models are more suitable for `chatting' with people, and they have also attracted attention from other authors \cite{li2015diversity,sordoni2015neural,shang2015neural,serban2016building}. Chen et al. \cite{xing2017topic} introduced the concept of LDA-based topic attention in generative chatbots. Our main goal is to construct an NMF-based, topic-aware generative chatbot . In this section, we provide a brief introduction to the key concepts such as the RNN encoder-decoder structure, attention, and nonnegative matrix factorization.

\subsection{RNN encoder-decoder}
\label{subsection:encoder_decoder}

The problem of constructing a chatbot can be formulated as follows. Let $\Omega$ be a finite set of words, and let $\Omega^{\ell}=\Omega\times \cdots \times \Omega$ be the set of sentences of length $\ell$ consisting of words from $\Omega$\footnote{Typical parameter choices are e.g., 10,000 common english words and $\ell=25$ for maximum word count in a sentence.}.

Sentences of varying lengths can be represented as elements of the set $\Omega^{\ell}$ by appending empty word tokens to the end of shorter sentences until the desired length is reached. Next, let $\mathcal{D}\subseteq \Omega^{\ell}\times \Omega^{\ell}$ be a large corpus of sentences that consists of question-answer pairs (e.g., from movie dialogues or Reddit threads). This will induce a probability distribution $p$ on $\Omega^{\ell}\times \Omega^{\ell}$ by 
\begin{align}
&p(q_{1},\dots,q_{\ell},a_{1},\dots,a_{\ell}) \\
&= \frac{1}{|\mathcal{D}|}
\begin{pmatrix}
\text{ of times that the question $(q_{1},\dots,q_{\ell})$} \\
\text{and answer $(a_{1},\dots,a_{\ell})$ appears in $\mathcal{D}$ }
\end{pmatrix}
\end{align}
The generative model's approach for sequence learning is to learn the best approximation $\hat{p}$ of the true joint distribution $p$. Once we have $\hat{p}$, we can build a chatbot that generates an answer $\mathbf{a}=(a_{1},\dots,a_{\ell})$ for a question $\mathbf{q}=(q_{1},\dots, q_{\ell})$ from the approximate conditional distribution $\hat{p}(\cdot \,|\, \mathbf{q})$. 

The encoder-decoder framework was first proposed in \cite{cho2014learning, sutskever2014sequence} in order to address the above problem, especially for machine translation.  The encoder uses an RNN to encode a given question $\mathbf{q}=(q_{1},\dots,q_{\ell})$ into a context vector $\mathbf{c}$, 
\begin{align}
  \mathbf{c} = f_{\textsf{c}}^{\textup{enc}}(\mathbf{h}_{1},\dots,\mathbf{h}_{\ell}), 
\end{align} 
where $(\mathbf{h}_{1},\dots,\mathbf{h}_{\ell})$ is a hidden state computed recursively from the input question $\mathbf{q}$ as 
\begin{align}
    \mathbf{h}_{i} = f_{\textsf{h}}^{\textup{enc}}(q_{i},\mathbf{h}_{i-1}).
\end{align}
Here $f_{\textsf{c}}^{\textup{enc}}$ and $f_{\textsf{h}}^{\textup{enc}}$ are nonlinear functions for computing encoder context vector and hidden states. For instance, in \cite{sutskever2014sequence}, $f_{\textsf{h}}^{\textup{enc}}$ and $f_{\textsf{c}}^{\textup{enc}}$ were taken to be Long-Short Term Memory (LSTM) \cite{hochreiter1997long} and the last hidden state, respectively. 

The decoder then generates a sequence of marginal probability distributions $(\hat{\mathbf{p}}_{1},\dots,\hat{\mathbf{p}}_{\ell})$ over $\Omega$ so that $\hat{\mathbf{p}}_{i}$ approximates the conditional distribution of the $i^{\text{th}}$ word in the answer given the question $\mathbf{q}$ and all previous predictions. By writing the joint probability distribution as a product of ordered conditionals, the decoder then computes a conditional probability distribution $\hat{p}(\cdot\,|\, \mathbf{q})$ on $\Omega^{\ell}$ for possible answers to the input question $\mathbf{q}$ by 
\begin{align}
    \hat{p}(a_{1},\dots,a_{\ell}\,|\, \mathbf{q}) &= \prod_{i=1}^{\ell} \hat{\mathbf{p}}_{i}(a_{i}) \\
    &\approx \prod_{i=1}^{\ell} p(a_{i}\,|\, a_{i-1},\dots,a_{1},\mathbf{q})\\
    &= p(a_{1},\dots,a_{\ell}\,|\, \mathbf{q}).
\end{align}

For computation, the decoder uses another RNN that computes prediction vectors $\hat{\mathbf{p}}_{i}$ recursively as 
\begin{align}
    \hat{\mathbf{p}}_{i} =  f_{\textsf{p}}^{\textup{dec}}(\hat{\mathbf{s}}_{i-1},\dots,\hat{\mathbf{s}}_{1},\mathbf{c}),
\end{align}
where $(\mathbf{s}_{1},\dots,\mathbf{s}_{\ell})$ are decoder hidden states, again computed recursively, this time as 
\begin{align}
\mathbf{s}_{i} = f_{\textsf{h}}^{\textup{dec}}(\hat{\mathbf{p}}_{i-1},\mathbf{s}_{i-1}, \mathbf{c}).
\end{align}
Here $f_{\textsf{p}}^{\textup{dec}}$ and $f_{\textsf{h}}^{\textup{dec}}$ are nonlinear functions for computing decoder prediction vector and hidden states.

The parameters in the nonlinear functions in $f_{\textsf{h}}^{\textup{enc}},f_{\textsf{c}}^{\textup{enc}},f_{\textsf{h}}^{\textup{dec}},f_{\textsf{p}}^{\textup{dec}}$ are then learned so that the following \textit{Kullback–Leibler divergence} \cite{kullback1951information} is minimized: 
\begin{align}\label{def:KL_divergence}
D(p \,\Vert\, \hat{p}) = -\sum_{(\mathbf{q},\mathbf{a})\in \Omega^{\ell}\times \Omega^{\ell}}p(\mathbf{q},\mathbf{a}) \log \left( \frac{\hat{p}(\mathbf{q},\mathbf{a})}{p(\mathbf{q},\mathbf{a})} \right). 
\end{align}
A typical technique for numerically solving the above optimization problem for training the encoder-decoder is called \textit{backpropagation through time} (BPTT), which is the usual method of backpropagation in deep feedforward neural networks applied to the `unfolded diagram' of RNN (see, e.g., \cite{robinson1987utility, werbos1988generalization, chauvin2013backpropagation}). However, as we deal with longer-range time dependencies (e.g., chatbots or machine translation), repeated multiplication causes the gradient to vanish at an exponential rate as it backpropogates through the network, which may cause deadlock in training phase. Some advanced versions of BPTT to handle this issue of the vanishing gradient problem have been developed, such as truncated BPTT with long short-term memory (LSTM) \cite{hochreiter1997long} and gated recurrent units (GRU) \cite{Chung2014gru}.

\subsection{The attention mechanism}

In the encoder-decoder framework, the input question $\mathbf{q}$ is mapped into a single context vector $\mathbf{c}$ by the encoder, which then is used as the initial hidden state for the decoder to generate prediction vectors $\hat{\mathbf{p}}_{1}, \dots, \hat{\mathbf{p}}_{\ell}$. However, limiting the decoder to just a single context vector might not be the best idea. What if instead, the decoder could utilize all encoder hidden states $\mathbf{h}_{1},\dots, \mathbf{h}_{\ell}$ and apply different weights to each of these states for each step of the decoder? This is the intuition behind the attention mechanism, and the hope is that it allows the decoder to `focus' on different features of the input as it creates each piece of the output.

The \textit{attention mechanism}, first proposed for image processing, was introduced to the field of natural language processing in 2014 \cite{P.J.Burt1988, bahdanau2014neural, MnihHGK14}. Single RNN models use hidden states to extract information to generate responses in a conversational setting. However, one problem with this structure is that sentences may be too long to be represented by a fixed-length context vector \cite{Weston2014, bahdanau2014neural}. Simultaneously, attention has proven to be effective and efficient in generation tasks, since it assists the network in focusing on specific parts of past information. Many methods have been suggested, including content-based attention \cite{graves2014neural}, additive attention \cite{bahdanau2014neural}, multiplicative attention \cite{Wang2017}, located based attention \cite{Luong2015}, and scaled attention \cite{vaswani2017attention}.

There has been significant work done with attention mechanisms in RNN-based language models; the most relevant to our work is by Xing et al. \cite{xing2017topic}. By using context-based attention as well as information from past input and predictions, Xing et al. implemented a \textit{topic attention} mechanism into the encoder-decoder model for topic-aware response generation. Their implementation utilizes the pre-trained Twitter LDA model \cite{zhao2011comparing} to acquire topical words from the input. This model assumes that each input corresponds to one topic within the LDA model, and each word in the input is either non-topical, or is a topic word corresponding to the current topic. The topic of the input question is then used as an input to the attention mechanism, which then biases the decoder to produce the predictions $\hat{\mathbf{p}}_{i}$ that have preferences to the given topic's words.





\subsection{Nonnegative matrix factorization}

\textit{Nonnegative matrix factorization} (NMF) is an algorithm for decomposing a nonnegative matrix into two smaller matrices whose product approximates the original matrix. NMF has been recognized as an indispensable  tool in text analysis, image reconstruction, medical imaging, bioinformatics, and many other scientific fields \cite{sitek2002correction, berry2005email, berry2007algorithms, chen2011phoenix, taslaman2012framework, boutchko2015clustering, ren2018non}. It is often used to extract features, or topics, from textual data. It aims to factor a given data matrix into nonnegative dictionary and code matrices in order to extract important topics:
\begin{align}
\texttt{Data}\approx \texttt{Dictionary}\times \texttt{Code}.
\end{align}
More precisely, one seeks to factorize a data matrix $X\in \mathbb{R}^{d\times n}_{\ge 0}$ into a product of low-rank nonnegative dictionary $W\in \mathbb{R}^{d\times r}$ and code $H\in \mathbb{R}^{r\times n}$ matrices by solving the following optimization problem
\begin{align}\label{eq:NMF_error1}
\inf_{W\in \mathbb{R}^{d\times r}_{\ge 0},\, H\in \mathbb{R}^{r\times n}_{\ge 0} }  \lVert X - WH  \rVert_{F}^{2},
\end{align}
where $\lVert A \rVert_{F}^{2} = \sum_{i,j} A_{ij}^{2}$ denotes the matrix Frobenius norm \cite{lee1999learning, lee2009semi}.

Once we have the above factorization, we can represent each column of the data matrix as a nonnegative linear combination of dictionary column vectors, where the coefficients are given by the corresponding column of the code matrix. Due to the nonnegativity constraint, each column of the data matrix is then represented as a nonnegative linear combination of dictionary atoms. Hence the dictionaries must be ``positive parts'' of the columns of the data matrix.

Many efficient iterative algorithms for NMF are based on block optimization schemes that have been proposed and studied, including the well-known multiplicative update method by Lee and Seung \cite{lee2001algorithms} (see \cite{gillis2014and} for a survey). To make the topics more localized and reduce the overlaps between the topics, one can also enforce sparseness on the code matrix \cite{Hoyer2004} in the optimization problem (\ref{eq:NMF_error1}). Moreover, algorithms for NMF have been extended to the online setting, where one seeks to learn dictionary atoms progressively from an input stream of data \cite{mairal2010online, guan2012online, zhao2016online}. Rigorous convergence guarantees for online NMF algorithms have been obtained in \cite{mairal2010online} for independent and identically distributed input data. Recently, these works were extended convergence guarantees of online NMF agorithms to the Markovian setting \cite{lyu2019online}, ensuring further versatility of NMF based topic modeling from input sequences generated by Markov Chain Monte Carlo algorithms. Furthermore, NMF-based approaches can be extended to the settings of dynamic \cite{Kasiviswanathan2011, saha2012learning}, hierarchical \cite{cichocki2007hierarchical}, and tensor factorization \cite{xu2013block, shashua2005non}.



\section{The RNN-NMF chatbot architecture}
\begin{figure*}[ht]
	\centering
	\includegraphics[width=1 \linewidth]{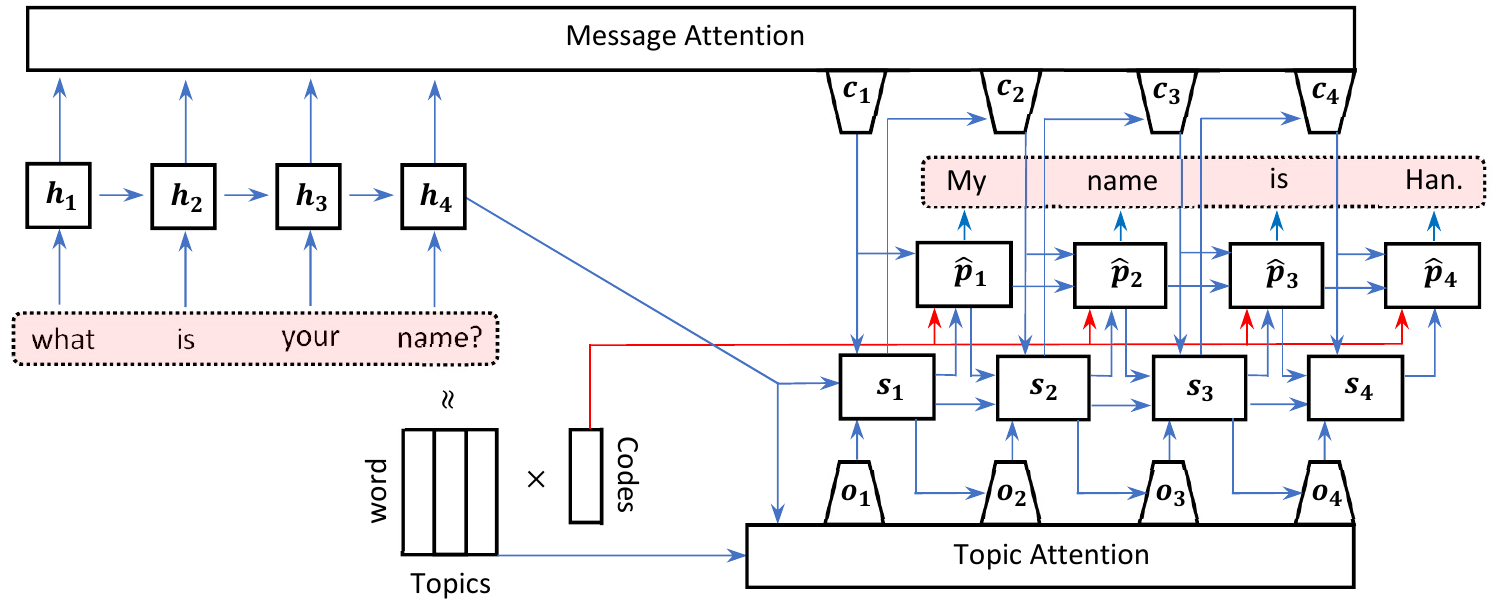}
	\caption{Illustration of the RNN-NMF chatbot structure. Input question ``What is your name?'' is fed into the encoder and hidden states $\mathbf{h}_{1},\dots,\mathbf{h}_{4}$ are computed recursively, which become basis for the message attention layer. On the other hand, the NMF topic representation of the input question is obtained using the pre-learned dictionary matrix, whose columns together with the last hidden state $\mathbf{h}_{4}$ become the basis for the topic attention layer. Decoder hidden states $\mathbf{s}_{1},\dots,\mathbf{s}_{4}$ as well as prediction vectors $\hat{\mathbf{p}}_{1},\dots,\hat{\mathbf{p}}_{4}$ are then computed recursively, making use of both message and topic attention vectors. When generating the output answer ``My name is Han.'' from the prediction vectors, the NMF-code of the input question is used to bias generation probability toward topic words.}\label{ChatbotStrucute}
\end{figure*}

In this section, we describe the architecture of our RNN-NMF chatbot. Our chatbot is based on the Encoder-Decoder structure that we described in Subsection \ref{subsection:encoder_decoder} with two additional message and topic attention layers, illustrated in Figure \ref{ChatbotStrucute}. Our main contribution is incorporating NMF into the topic attention layer, which provides a simple and efficient way to make the encoder-decoder based chatbot topic-aware.

\subsection{Obtaining topic representation with NMF}
\label{subsection:NMF_coding}
The algorithm begins with two data sets---a corpus $\mathcal{D}$ of question-answer pairs of sentences and a corpus $\mathcal{S}$ of documents for topic modeling. In order to learn topic vectors from $\mathcal{S}$, we first turn it into a $\texttt{Word}\times \texttt{Doc}$ matrix by representing each document as a \texttt{Word}-dimensional column vector using a bag-of-words representation \cite{harris1954distributional}. We use  standard (or online) NMF algorithms (e.g., \cite{lee1999learning, Mairal2010, lyu2019online}) to obtain the following factorization 
\begin{align*}
    (\texttt{Word}\times \texttt{Doc}) \approx (\texttt{Word}\times \texttt{Topic}) \times (\texttt{Topic}\times \texttt{Doc}).
\end{align*} 
We will denote the dictionary matrix $(\texttt{Word}\times \texttt{Topic})$ by $W=(\mathbf{w}_{1},\dots,\mathbf{w}_{r})$, where each $\mathbf{w}_{i}$ denotes the $i^{\text{th}}$ column vector of $W$.

Now for each question-answer pair $(\mathbf{q},\mathbf{a})$ of column vectors in the corpus $\mathcal{D}$, we obtain the NMF \textit{code} of question $\mathbf{q}$, which we denote by $\texttt{Code}(\mathbf{q})$, so that we have the following approximate \textit{topic representation}  of the input question:
\begin{align}\label{eq:topic_representation}
\mathbf{q} \approx W \,\texttt{Code}(\mathbf{q}).
\end{align}
A standard approach for finding this is solving the convex optimization problem
\begin{align}\label{eq:code_optimization}
    \texttt{Code}(\mathbf{q}) = \textup{argmin}_{\mathbf{k}} \lVert \mathbf{q} - W \mathbf{k}\rVert_{F}^{2} + \lambda \lVert \mathbf{k} \rVert_{1},
\end{align}
where $\lambda>0$ is a fixed $L_{1}$ regularization parameter. This can be computed using a number of well-known algorithms (e.g., LARS \cite{efron2004least}, LASSO \cite{tibshirani1996regression}, and feature-sign search \cite{lee2007efficient}). 

However, in order to see a diverse set of words within each column, the topic matrix $W$ is typically learned from a large text corpus matrix whose columns are bag-of-words representations of documents, not sentences. In this case, the question vectors $\mathbf{q}$ are too sparse to be represented as nonnegative linear combinations of the dense topic vectors in $W$, so the sparse coding problem \eqref{eq:code_optimization} will result in (nearly) all zero solutions for any question.

In order to overcome the above issue, we propose an alternative formula to compute $\texttt{Code}(\mathbf{q})$:
\begin{align}
    \texttt{Code}(\mathbf{q}) = W^{T} \mathbf{q}.
\end{align}
Namely, the contribution of the $j^{\text{th}}$ topic vector $\mathbf{w}_{j}$ in the question $\mathbf{q}$ is computed as the inner product $\mathbf{w}_{j}^{T}\mathbf{q}$, which measures the correlation between the question and the corresponding topic vector. Hence the topic vectors obtained by NMF will act as a `field of correlated words', and the coefficients in $W^{T}\mathbf{q}$ will tell us how much the chatbot should use each field (or topic) in generating a response. (See the last eq. in \eqref{eq:predicted_dist} below.)

\subsection{Encoder} 
\label{subsection:encoder}

Fix a question-answer pair $(\mathbf{q},\mathbf{a})$ from the corpus $\mathcal{D}$ and write $\mathbf{q}=(q_{1},\dots,q_{\ell})$. The encoder uses an RNN to encode a question $\mathbf{q}$ as a sequence of hidden states $(\mathbf{h}_{1},\dots,\mathbf{h}_{\ell})$ recursively as 
\begin{align}
  \mathbf{h}_{i} = f_{\textsf{h}}^{\textup{enc}}(q_{i},\mathbf{h}_{i-1}),
\end{align}
where we use the nonlinear function $f_{\textsf{h}}^{\textup{enc}}$ to be the gated recurrent unit (GRU) \cite{Chung2014gru}. Namely for parameter matrices $W_{z},\,W_{r},\,W_{s},\,U_{z},\,U_{r},\,U_{s}$, and sigmoid function $\sigma$, we define 
\begin{align}
    \begin{cases}
        \mathbf{z} = \sigma(W_{z}\mathbf{x}_{t} + U_{z}\mathbf{h}_{t-1}) \\
        \mathbf{r} = \sigma(W_{r}\mathbf{x}_{t} + U_{r}\mathbf{h}_{t-1}) \\
        \mathbf{s}=\tanh(W_{s}\mathbf{x}_{t} + U_{s}(\mathbf{h}_{t-1} \mathbf{r}^{T}) ) \\
         \mathbf{h}_{t} = (\mathbf{1}-\mathbf{z}) \mathbf{s}^{T} + \mathbf{z} \mathbf{h}_{t-1}^{T}.
    \end{cases}
\end{align}
In addition, the Encoder finds the NMF code $\mathtt{Code}(\mathbf{q})=(k_{1},\dots,k_{r})^{T}$ of $\mathbf{q}$ by solving \eqref{eq:code_optimization}.  

\subsection{Decoder: message and topic attention}
\label{subsection:decoder}

Once the question-answer pair $(\mathbf{q},\mathbf{a})$ is run through the encoder, we have the encoder hidden states $(\mathbf{h}_{1},\dots,\mathbf{h}_{\ell})$ and the NMF-code $\mathtt{Code}(\mathbf{q})={(k_{1},\dots, k_{r})}^{T}$. These are passed as inputs to the decoder. Using $\ell'$ to denote the maximum length of the output sentence, the decoder outputs a sequence $(\hat{\mathbf{p}}_{1},\dots,\hat{\mathbf{p}}_{\ell'})$ of prediction vectors using GRU and message and topic attention. Let $\mathbf{s}_{1},\dots,\mathbf{s}_{\ell'}$ denote decoder hidden state, where we set $\mathbf{s}_{0}=\mathbf{h}_{\ell}$, the last encoder hidden state.

We will now describe how to update the state of the decoder. First, suppose the vectors $\hat{\mathbf{p}}_{t-1},\, \mathbf{s}_{t-1},\, \mathbf{c}_{t},\,\mathbf{o}_{t}$ have been computed by the decoder. The next decoder hidden state $\mathbf{s}_{t}$ is computed by 
\begin{align*}
\mathbf{s}_{t} &= f_{\textsf{h}}^{\textup{dec}}(\hat{\mathbf{p}}_{t-1}, \mathbf{s}_{t-1}, \mathbf{c}_{t},\mathbf{o}_{t}) \\
&= \sigma(W_{p}^{(s)} \hat{\mathbf{p}}_{t-1} + W_{s}^{(s)}  \mathbf{s}_{t-1} + W_{c}^{(s)}\mathbf{c}_{t} + W_{o}^{(s)} \mathbf{o}_{t} + \mathbf{b}^{(s)})
\end{align*}
for sigmoid function $\sigma$, nonlinear function $f_{\textsf{h}}^{\textup{dec}}$, and suitable parameters $W_{p}^{(s)},\, W_{s}^{(s)},\, W_{c}^{(s)},\, W_{o}^{(s)},$ and $\mathbf{b}^{(s)}$. The superscripts of the parameters denotes what they are used to calculate, and the subscripts denotes to what they are applied.

Next, suppose a decoder hidden state $\mathbf{s}_{t-1}$ is given. Then three vectors are computed by the decoder: message attention $\mathbf{c}_{t}$, topic attention $\mathbf{o}_{t}$, and predicted distribution $\mathbf{\hat{p}}_{t}$. The message attention $\textbf{c}_{t}$ is defined via the following equations:
\begin{align}
    \begin{cases}
        \xi^{(c)}_{tj} &= \eta^{(c)} (\mathbf{s}_{t-1},\mathbf{h}_j) \\
        \alpha_{tj}^{(c)} &= {\Big(\sum_{k=1}^{\ell}\exp\big({\xi^{(c)}_{tk}}\big)\Big)}^{-1}\exp\big(\xi^{(c)}_{tj}\big)  \\
        \mathbf{c}_{t} &= \sum_{j=1}^{\ell}\alpha_{tj}^{(c)}\mathbf{h}_j,
    \end{cases}
\end{align}
where $\eta^{(c)}$ is a multi-layer perceptron with $\tanh$ as an activation function, $t$ indexes the decoder hidden state, $j$ indexes the encoder hidden state, and the superscript $(c)$ denotes that these values relate to the message attention $\mathbf{c}_{t}$. In words, the message attention $\mathbf{c}_{t}$ is a linear combination of the encoder hidden states $\mathbf{h}_{1},\dots,\mathbf{h}_{\ell}$, where the `importance' $\alpha^{(c)}_{tj}$ of the $j^{\text{th}}$ encoder hidden state $\mathbf{h}_{j}$ depends on $\mathbf{h}_{j}$ itself as well as the previous decoder hidden state $\mathbf{s}_{t-1}$.

Similarly, the topic attention $\textbf{o}_{t}$ is computed via the following equations:
\begin{align}
\begin{cases}
\xi^{(o)}_{tj} &= \eta^{(o)}(\mathbf{s}_{t-1},\mathbf{w}_{j},\mathbf{h}_\ell) \\
\alpha_{tj}^{(o)} &= {\Big(\sum_{k=1}^{r}\exp\big({\xi^{(o)}_{tk}}\big)\Big)}^{-1}\exp\big(\xi^{(o)}_{tj}\big)  \\
\mathbf{o}_{t} &= \sum_{j=1}^{r}\alpha_{tj}^{(o)}\mathbf{w}_j,
\end{cases}
\end{align} 
where $\eta^{(o)}$ is a similar perceptron as $\eta^{(c)}$, $t$ indexes the decoder hidden states, $j$ indexes the $r$ columns of the dictionary matrix $W$ learned by NMF from the corpus $\mathcal{S}$ (see Subsection \ref{subsection:NMF_coding}), and the superscript $(o)$ denote that these values relate to the topic attention $\mathbf{o}_{t}$. Notice that we only use the last encoder hidden state $\mathbf{h}_{\ell}$ in computing topic attention. This is because the `topic' of the input question $\mathbf{q}$ is information about the entire sentence, which should be encoded in the last hidden state $\mathbf{h}_{\ell}$. However, we do use all topic vectors $\mathbf{w}_{1},\dots,\mathbf{w}_{r}$ in calculating the topic attention.

\subsection{Decoder: predicted distribution with NMF-biasing}
It remains to describe how the Decoder computes the predicted distribution $\hat{\mathbf{p}}_{t}$ given the decoder hidden state $\mathbf{s}_{t}$, which  is a probability distribution over the set $\Omega$ of all possible vocabularies that is suppose to be a good approximation of the true conditional probability $\mathbf{p}_{t}$ of the $t$th word in the correct response. 

Recall that the NMF-dictionary matrix $W=(\mathbf{w}_{1},\dots,\mathbf{w}_{r})$ is given. For each word $\omega\in \Omega$ and $1\le j \le r$, we write $\mathbf{1}(\omega\in \mathbf{w}_{j})$ for the indicator function which describes whether $\omega$ appears in $\mathbf{w}_{j}$. We also write $\mathbf{w}$ for the column indicator vector of the word $\omega$. We define the predicted distribution $\hat{\mathbf{p}}_{t}$ through the following equations:
\begin{align}\label{eq:predicted_dist}
  \begin{cases}
        \Psi^{(c)}_{t}(\omega) \\
        \qquad =  \sigma(\mathbf{w}^T(W_{s}^{(c)} \mathbf{s}_{t} +  W_{p}^{(c)} \hat{\mathbf{p}}_{t-1}+ W_{c}^{(c)} \mathbf{c}_{t}+\mathbf{b}^{(c)})) \\
        \Psi^{(o)}_{t}(\omega) \\
        \qquad =  \sigma(\mathbf{w}^T(W_{s}^{(o)} \mathbf{s}_{t} + W_{p}^{(o)} \hat{\mathbf{p}}_{t-1} + W_{o}^{(o)} \mathbf{o}_{t}+\mathbf{b}^{(o)})) \\
        \hat{\mathbf{p}}_{t}(\omega) \propto \exp(\Psi^{(c)}_{t}(\omega)) \\
        \qquad \qquad + \left(\sum_{j=1}^{r} k_{j} \mathbf{1}(\omega\in \mathbf{w}_{j}) \right) \exp(\Psi^{(o)}_{t}(\omega)),
      \end{cases}
\end{align}
where $W_{s}^{(c)},\, W_{p}^{(c)},\, W_{c}^{(c)},\, W_{s}^{(o)},\, W_{p}^{(o)},\, W_{o}^{(o)},\, \mathbf{b}^{(c)}$, and $\mathbf{b}^{(o)}$ are parameters with subscripts denoting which values they are applied to and superscripts on the $\Psi$ function that they relate to (with $(c)$ referring to the context attention and $(o)$ referring to the topic attention). In addition, $\propto$ means ``proportional to'' in the traditional sense (the last expression is not an equation because of the unknown normalization constant). 

In the traditional sequence-to-sequence model, the predicted distribution $\hat{\mathbf{p}}_{t}$ for the $t^{\text{th}}$ word in the response is proportional to the exponential of $\Psi^{(c)}$, which is the first term in the last expression in \eqref{eq:predicted_dist}. In our model, we give additional bias toward topic words according to the NMF-code $\mathtt{Code}(\mathbf{q})$ of the input sentence through the second term in the last equation in \eqref{eq:predicted_dist}. Recall that due to the approximate nonnegative factorization \eqref{eq:topic_representation}, we can view the $j^{\text{th}}$ entry $k_{j}$ of $\mathtt{Code}(\mathbf{q})$ as the importance of the $j^{\text{th}}$ topic vector $\mathbf{w}_{j}$ given by the NMF-topic matrix $W$. Hence if a word $\omega$ belongs to the $j^{\text{th}}$ topic vector $\mathbf{w}_{j}$, it gets extra non-negative bias proportional to the corresponding code $k_{j}$ as well as the previous decoder state vectors through the function $\Psi^{(o)}$.

\subsection{Training the model and generating response}
We train the parameters in the RNN-NMF chatbot by approximately solving the following optimization problem
\begin{align}\label{eq:model_training}
\texttt{parameters} = \textup{argmin} \sum_{(\mathbf{q},\mathbf{a})\in \mathcal{D}} \sum_{t=1}^{\ell'} D(\mathbf{p}_{t}\,\Vert\, \hat{\mathbf{p}}_{t}),
\end{align}
where $D(\cdot\,\Vert\,\cdot)$ denotes the KL-divergence defined in \eqref{def:KL_divergence}. Recall that $\mathbf{p}_{t}$ denotes the indicator vector representation (or Dirac delta mass) of the $t^{\text{th}}$ word $a_{t}$ in the correct answer $\mathbf{a}$, and $\hat{\mathbf{p}}_{t}$ denotes the predicted distribution produced by the RNN-NMF chatbot. Hence \eqref{eq:model_training} amounts to optimizing the parameters so that the KL-divergence \eqref{def:KL_divergence} between the true response and the approximate joint distributions on the question-answer pairs are minimized. For numerical computation, one may use backpropagation through time algorithms \cite{robinson1987utility, werbos1988generalization, chauvin2013backpropagation}. After the training, the chatbot generates the response $\hat{\mathbf{a}}=(\hat{a}_{1},\dots,\hat{a}_{\ell'})$ for a given question $\mathbf{q}$ by sequentially sampling $\hat{a}_{i}$ from the predicted distribution $\hat{\mathbf{p}}_{i}$ for $i=1,2,\dots, \ell'$.

By using Markov Chain Monte Carlo (MCMC) sampling, both for the training and the generation steps, the normalization constant of the predicted probability distribution $\hat{\mathbf{p}}_{i}$ does not have to be computed explicitly. This improves the speed of both steps of the algorithm especially when the vocabulary space $\Omega$ is large. A standard method of constructing a Markov chain $(X_{t})_{t\ge 0}$ is the Metropolis-Hastings algorithm (see., e.g., \cite{levin2017markov}). This method can be used to construct a chain of words in $\Omega$ so that the distribution of $X_{t}$ converges to $\hat{\mathbf{p}}_{i}$.

Given the word $X_{t}=\omega$ at iteration $t$, the next word $X_{t+1}$ is obtained as follows:
\begin{description}
    \item{(i)} Sample a word $\omega'\in \Omega$ according to the marginal distribution $p^{(i)}$ of the $i^{\text{th}}$ word in response from the joint distribution $p$ induced by the corpus $\mathcal{D}$.
    
    \item{(ii)} Denote $\psi(\cdot)$ for the right hand side of the last equation of \eqref{eq:predicted_dist}. Compute the acceptance probability 
    \begin{align}
        \lambda = \min\left( \frac{\psi(\omega')}{\psi(\omega)} \frac{p^{(i)}(\omega)}{p^{(i)}(\omega')},\,1  \right).
    \end{align}
    
    \item{(iii)} Sample $X_{t+1}=\omega'$ with probability $\lambda$ and $\omega$ with probability $1-\lambda$. 
\end{description} 

Once we have the above chain $(X_{t})_{t\ge 0}$ converging to $\hat{\mathbf{p}}_{i}$, we can run this chain for several steps to approximately sample the response word $a_{i}$ from $\hat{\mathbf{p}}_{i}$. During training, the empirical distribution of the chain gives an approximation of $\hat{\mathbf{p}}_{i}$, which we may use for approximately computing the KL-divergence $D(\mathbf{p}_{i}\,\Vert\, \hat{\mathbf{p}}_{i})$. 

\subsection{Dimension reduction by word embedding}
We will now discuss a technical point in formulating the language modeling task more efficiently by using a word embedding. 

In the most basic representation, we can represent each word $\omega$ as the indicator vector (or one-hot encoding) $\mathbf{1}(\omega)$ that has a one at the entry corresponding to  $\omega$ and zeros at all entries. However, this method is extremely memory-inefficient and it becomes a problem when the maximum sequence length and vocabulary size are large enough. In addition, the input layer of a neural network needs to be extremely large in order to have an input node that corresponds to every single word in the vocabulary. The number of weights to train in that first layer alone will be the product of the vocabulary size and the size of the first hidden layer. This makes training extremely slow.

A widely used approach to solve the above problem is to encode each word as a lower-dimensional vector with multiple non-zero entries as opposed to an indicator vector. This enables us to reduce the dimension of our word encoding. A good way of accomplishing this is to embed words in a space that preserves some of the relationships between different words. One widely-used word embedding is called GloVe, and it combines both global matrix factorization and local context window methods in order to produce a meaningful embedding \cite{Pennington2014}. This is what we will use below.

\section{Experiment}
In this section, we present the empirical results of our RNN-NMF chatbot model. We train three different topic-aware chatbots, each using their own topic matrices built from different datasets: 1) \texttt{DeltaAir} (Delta Airline customer support records) \cite{delta},  2) \texttt{Shakespeare} (all lines in all of Shakespeare's plays) \cite{shakespeare}, and 3) \texttt{20NewsGroups} (20 News groups articles) \cite{20newsgroups}. Using each of these topic matrices, we train our RNN-NMF chatbot on the Cornell Movie Dialogue dataset \cite{danescu2011cornell}. We also train an additional non-topic model that does not use a topic matrix and will serve as a baseline model for comparison.

\subsection{Obtaining topic matrices by NMF}

In order to learn topic matrices using NMF, we first convert each document in the corpus into a single string of words. Each string is then encoded into a bag-of-words model which we aggregate into a matrix where each column represents a document. A term frequency---inverse document frequency transformer \cite{leskovec2014mining} is then applied to this matrix to give more significant words more weight. A standard NMF algorithm is then used to learn 10 topic vectors.

Below we list the top 10 highest-weighted words in the NMF dictionaries for arbitrarily chosen five of the topics for each of the three data sets.

\begin{description}
    \item{\textbf{Topics learned from DeltaAir}}
    \item{Topic \#1}: thank, welcome, flying, feedback, appreciate, great, delta, loyalty, sharing, day
    \item{Topic \#2}: flight, delayed, delay, sorry, time, crew, gate, hours, hi, hear
    \item{Topic \#3}: seat delta upgrade seats comfort middle available class hi economy
    \item{Topic \#4}: let, know, assistance, need, sorry, amv, rebooking, assist, delay, apologies
    \item{Topic \#5}: bag, baggage, check, claim, airport, bags, lost, checked, luggage, team
\end{description}

\vspace{0.1cm}
\begin{description}
    \item{\textbf{Topics learned from Shakespeare}}
    \item{Topic \#1}: enter, messenger, king, attendants, servant, gloucester, lords, duke, queen, henry
    \item{Topic \#2}: lord, good, ay, know, noble, say, tis, king, gracious, did
    \item{Topic \#3}: exit, servant, falstaff, messenger, gentleman, body, lucius, boy, gloucester, cassio
    \item{Topic \#4}: thy, hand, father, heart, hath, love, life, let, master, face
    \item{Topic \#5}: sir, good, know, ay, pray, john, man, say, did, marry
\end{description}

\vspace{0.1cm}
\begin{description}
    \item{\textbf{Topics learned from 20NewsGroups}}
    \item{Topic \#1}: card, video, monitor, drivers, cards, bus, vga, driver, color, ram
    \item{Topic \#2}: god, jesus, bible, christ, faith, believe, christian, christians, church, sin
    \item{Topic \#3}: game, team, year, games, season, players, play, hockey, win, player
    \item{Topic \#4}: car, new, 00, sale, 10, price, offer, condition, shipping, 20
    \item{Topic \#5}: edu, soon, cs, university, com, email, internet, article, ftp, send
\end{description}

\begin{table*}[htbp]
	\centering
	\begin{tabular}{|c|c|c|c|c|}
		\hline
		& \texttt{Non-topic}  & \texttt{DeltaAir} & \texttt{Shakespeare} & \texttt{20NewsGroups} \\
		\hline
		Training loss & 0.2847  & 0.1937 &  0.1708  &  0.2061 \\
		\hline
	\end{tabular}
	\vspace{0.3cm}
	\caption{Training loss of the four chatbot models trained on the Cornell Movie Dialogue dataset. Lower training losss is better and indicates that the model has learned the desired joint distribution on the question-answer pairs.}
	\label{table:perplexity}
\end{table*}

\begin{table*}[ht]
	\centering
	\includegraphics[width=1 \linewidth]{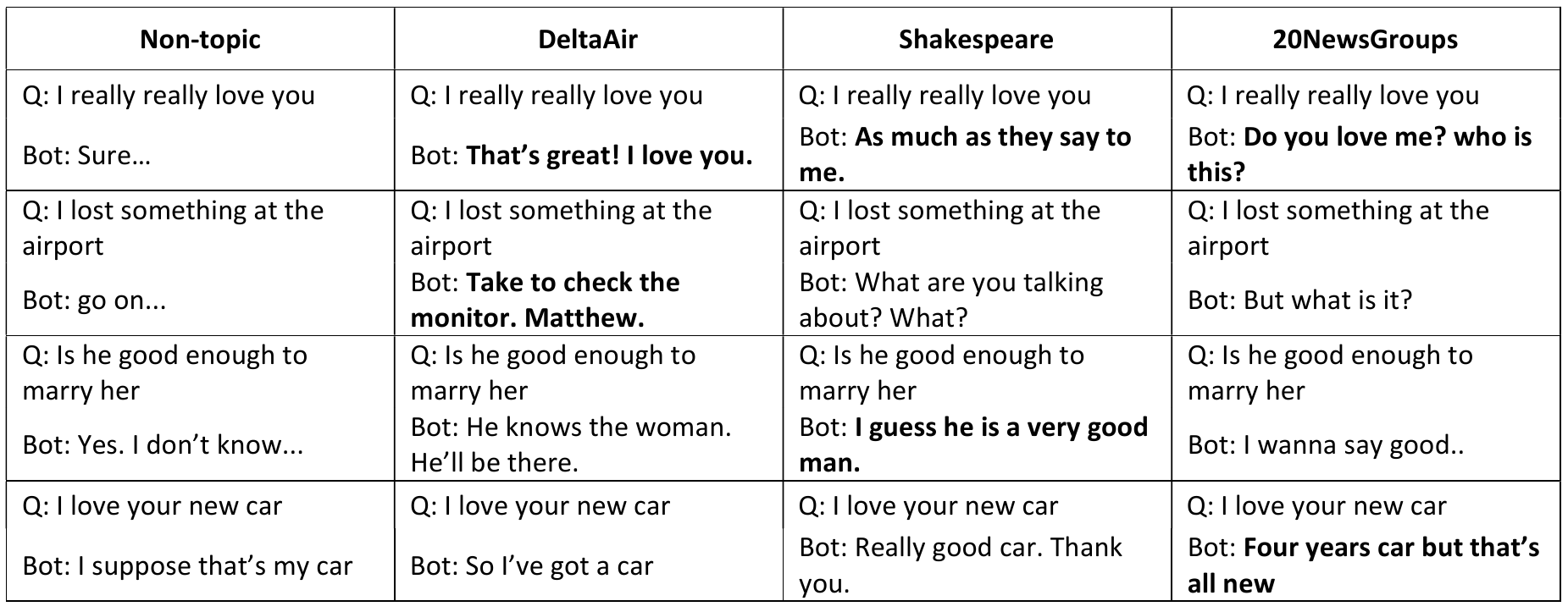}
	\vspace{-0.3cm}
	\caption{ Examples of conversation with chatbots with different NMF-topic filters. A bold response by the chatbot indicates that we find it to be a reasonable, non-generic response to the input.}
	\label{table:chat_ex}
\end{table*}

\subsection{Training the RNN-NMF chatbot}

With the three topic matrices obtained by NMF in the previous subsection, we now train our RNN-NMF chatbot on the Cornell Movie Dialogue data set, which is a classical data set for training RNN based chatbot. We note that training on a much larger data set such as Reddit or Twitter data would likely improve the performance of the chatbot, but training on the smaller Cornell data set suffices for our goal of comparing the performance of non-topic and various topic versions of the chatbot. Moreover, in order to speed up the training process, we restrict our model to the 18,000 most commonly used English words and mask all the other words by replacing them with a special token that denotes unknown words.

Our training uses a single NVIDIA GeForce GTX 1660 Ti GPU. We use the \texttt{Adagrad} optimizer, the batch size is 64, the dropout rate is 0.1, the hidden size for the word embedding is 500, and the total number of  training iterations is 64,000 (87 epochs). We also use teacher forcing \cite{williams1989learning} to accelerate the training, with a teacher forcing rate of 1.

\subsection{Comparison and analysis}

In this subsection, we present output from the four chatbot models (\texttt{Non-topic}, \texttt{DeltaAir}, \texttt{Shakespeare}, and \texttt{20NewsGroups}). Recall that all chatbot models have been trained on the same Cornell Movie Dialogue, but the use of different topic matrices influence the training via topic attention and topic-biasing in computing the predicted probability distribution  \eqref{eq:predicted_dist}. In some sense, each topic matrix learned by NMF from different data sets provide a priori `filter' or `bias' in choosing words in response. In this sense, we will refer to each topic matrix used in training the chatbot as its \textit{NMF-topic filter}.

For a quantitative comparison, in Table \ref{table:perplexity} we provide the training loss (i.e., the KL-divergence between the true and predicted distribution in \eqref{def:KL_divergence}) of each of the four chatbots at the end of training. Lower training loss indicates that the model has learned the desired joint distribution on the question and answer pairs from the Cornell data set. Indeed, training loss of all of the three topic-aware chatbot models are comparably lower than that of the non-topic model.

For qualitative comparison, we give examples of conversation with the four models in Table \ref{table:chat_ex}. We observe that in general, topic-aware chatbots give more non-generic and context-dependent responses to various input questions. Moreover, when we ask questions closely related to each topic matrix (\texttt{DeltaAir}, \texttt{Shakespeare}, and \texttt{20NewsGroups}), we observe that the chatbot with the corresponding NMF-topic filter gives the most appropriate answer using relevant topic words more frequently. Namely, for the question ``I lost something at the airport'', the words ``lost'' and ``airport'' belong to Topic \#5 of the \texttt{DeltaAir} data and the related topic word ``check'' does appear in the corresponding chatbot's response. Also, for the question, ``Is he good enough to marry her'', ``good'' and ``marry'' belong to \texttt{Shakespeaere}'s Topic \#5, and the corresponding chatbot's answer contains associated topic words ``good'' and ``man''. Lastly, for the question ``I love your new car'', the words ``car'' and ``new'' belong to Topic \#4 of the \texttt{20NewsGroups} data, and both of them as well as additional topic word ``years'' from the same topic vector appear in the corresponding chatbot's response.

We will now discuss the topic-awareness of the chatbot in relation to the conversation and topic vocabularies. Recall that we train our RNN-NMF chatbot model on the Cornell Movie Dialogue with masking all but the 18,000 most commonly used English words. Some of the topic words that we learn by NMF from different data sets (e.g., `ay', `thy', and `exeunt' from \texttt{Shakespeare}) are not necessarily contained in this list of 18,000 words or frequently appear in the movie dialogue. Since the training minimizes the divergence between the predicted and true output probability distributions, any response using topic words that do not appear in the actual responses in the conversation data set will be suppressed during the training process. Indeed, the responses in Table \ref{table:chat_ex} all consist of very generic and commonly used words. Hence, if we train our RNN-NMF chatbot model on a large enough conversation data that contains most of the topic words, the chatbots will likely give more topic-oriented responses. As this requires more powerful computational resources, we leave it for our future work.

\section{Concluding remarks and future works}

In this paper, we have constructed a model for a topic-aware chatbot by combining several existing text-generation models that utilize RNN structures with NMF-based topic modeling. Before we train the chatbot on a conversation data set, a topic matrix is obtained by NMF from a separate data set different kinds. For a given input question, its correlation with the learned topic vectors is computed and then fed into the decoder so that it learns to prefer relevant topic words over more generic ones. We demonstrated our RNN-NMF chatbot architecture using four variants: \texttt{Non-topic}, \texttt{DeltaAir}, \texttt{Shakespeare}, and \texttt{20NewsGroups}. Not only do the topic-aware chatbots achieve comparably lower training loss on the data set than the non-topic one, they qualitatively and contextually give the most relevant answer depending on the topic of question. 

Our work serves as the first proof of concept for using NMF-based topic modeling to create a topic-aware, RNN-based chatbot. This opens up a number of future research possibilities, which build on recent developments in the literature of NMF. Below we discuss three possibilities.

\subsection{Learning topics from MCMC trajectories.}
Suppose we have to learn the topic vectors from documents sampled using Markov chain Monte Carlo methods. For instance, say we would like to make our chatbot  aware of trending topics from Twitter or the internet in general. Since it is not possible to collect or process the entirety of Twitter or the internet, we may use a Markov chain-based sampling algorithm to obtain a sequence of samples of the texts and tweets (e.g., Google's PageRank or random walk based exploration algorithm). Recent work guarantees convergence of an online NMF algorithm on a Markovian sequence of data \cite{lyu2019online}. Therefore, our current architecture can easily be extended to such a setting. Namely, we learn the NMF-topic matrix by using the Markovian NMF algorithm in \cite{lyu2019online} from an MCMC trajectory exploring the text sample space. We can then proceed by training our chatbot with this learned topic matrix.

\subsection{Transfer learning.}
In our RNN-NMF chatbot architecture, the use of the NMF-topic filter is hard-coded in the sense that the NMF topic matrix of choice is used in the training step. Hence, if we wanted to apply a different topic filter to the chatbot, we would need to train the entire model from the beginning. However, if we were able to use an existing general language model instead of the RNN structure, we would be able to switch topic filters by merely fine-tuning the parameters of the NMF-topic layer. One possiblilty is to use BERT \cite{paulin2015concentration} as the general language model, and add a NMF-topic later so that the predicted probability distribution is computed in a similar way to \eqref{eq:predicted_dist}.

\subsection{Dynamic topic-awareness.}
One of the main advantages in using NMF-based topic modeling instead of LDA is that the topic matrix can be very easily learned from a new or even a time-varying data set. Hence we can make our RNN-NMF chatbot architecture `online', so that the underlying online NMF algorithms (e.g., \cite{Mairal2010, lyu2019online}) can constantly learn new topics from a time-varying data set (e.g., collective chat history of users). Then, the RNN can retrain the parameters in the topic-layer on the fly in order to adapt to current topics.

\section*{Acknowledgment}

This work has been partially supported by NSF DMS grant 1659676, NSF CAREER DMS \#1348721, and NSF BIGDATA \#1740325. The authors are grateful for Andrea Bertozzi for the generous support and Blake Hunter for helpful and inspiring discussions.

\end{multicols}

\vspace{0.3cm}

\small{
    \bibliographystyle{amsalpha}
    \bibliography{finalbib}
}

\end{document}